\documentclass[sigconf, screen, authorversion, nonacm, natbib=false]{acmart}

\AtBeginDocument{%
  \providecommand\BibTeX{{%
    \normalfont B\kern-0.5em{\scshape i\kern-0.25em b}\kern-0.8em\TeX}}}

\newcommand{\bfbar}[1]{\mathbf{\overline{#1}}}
\newcommand{\bftilde}[1]{\mathbf{\tilde{#1}}}

\newcommand{\bfa}[0]{\mathbf{a}}
\newcommand{\bfb}[0]{\mathbf{b}}
\newcommand{\bfc}[0]{\mathbf{c}}
\newcommand{\bfu}[0]{\mathbf{u}}
\newcommand{\bfx}[0]{\mathbf{x}}
\newcommand{\bfy}[0]{\mathbf{y}}
\newcommand{\bfv}[0]{\mathbf{v}}
\newcommand{\bfz}[0]{\mathbf{z}}
\newcommand{\bfW}[0]{\mathbf{W}}
\newcommand{\bfLambda}[0]{\mathbf{\Lambda}}

\newcommand{\bfA}[0]{\mathbf{A}}
\newcommand{\bfP}[0]{\mathbf{P}}
\newcommand{\bfB}[0]{\mathbf{B}}
\newcommand{\bfC}[0]{\mathbf{C}}
\newcommand{\bfD}[0]{\mathbf{D}}
\newcommand{\bfE}[0]{\mathbf{E}}

\newcommand{\bfPhi}[0]{\mathbf{\Phi}}
\newcommand{\bfTheta}[0]{\mathbf{\Theta}}

\newcommand{\bfdelta}[0]{\mathbf{\delta}}

\newcommand{\integrate}[4]{\int_{#3}^{#4} #1 \mathrm{d}#2}

\newcommand{\bfxdot}{\dot{\mathbf{x}}}

\newcommand{\R}[1]{\mathbb{R}^{#1}}
\newcommand{\C}[1]{\mathbb{C}^{#1}}

\newcommand{\bigorder}[1]{$\mathcal{O}\left( #1 \right)$}

\newcommand{\secref}[1]{sec.~\ref{sec:#1}}
\newcommand{\figref}[1]{fig.~\ref{fig:#1}}
\newcommand{\tabref}[1]{tab.~\ref{tab:#1}}

\newcommand{\Tabref}[1]{Tab.~\ref{tab:#1}}

\newcommand{\minisec}[1]{\par\smallskip\noindent\textbf{#1.}}
\usepackage{siunitx}
\usepackage{booktabs}
\usepackage{multirow}
\usepackage[utf8]{inputenc}
\usepackage{pifont} %

\DeclareSIUnit{\million}{\text{M}}

\setcopyright{acmlicensed}
\copyrightyear{2024}
\acmYear{2024}

\acmConference[ICONS'24]{International Conference on Neuromorphic Systems}{July 03-- August 02, 2024}{Arlington, Virginia}

\usepackage[
backend=biber,
style=acmauthoryear,
]{biblatex}
\begin{document}

\title{Scalable Event-by-event Processing of Neuromorphic Sensory Signals With Deep State-Space Models
}

\author{Mark Sch{\"o}ne}
\affiliation{%
  \institution{TU Dresden}
  \city{Dresden}
  \country{Germany}}
\orcid{0000-0003-0148-3764}
\email{mark.schoene@tu-dresden.de}
\author{Neeraj Mohan Sushma}
\affiliation{%
  \institution{Ruhr-University Bochum}
  \city{Bochum}
  \country{Germany}
}
\author{Jingyue Zhuge}
\affiliation{%
  \institution{TU Dresden}
  \city{Dresden}
  \country{Germany}}
\author{Christian Mayr}
\affiliation{%
    \institution{Center for Scalable Data Analytics and Artificial Intelligence (ScaDS.AI)}
    \city{}
    \country{}
}
\affiliation{%
    \institution{Centre for Tactile Internet with Human-in-the-Loop (CeTI)}
    \city{}
    \country{}
}
\affiliation{%
    \institution{TU Dresden}
    \city{Dresden}
    \country{Germany}
}
\author{Anand Subramoney}
\affiliation{%
    \institution{Royal Holloway, University of London}
    \city{Egham}
    \country{United Kingdom}
}
\author{David Kappel}
\affiliation{%
  \institution{Ruhr-University Bochum}
  \city{Bochum}
  \country{Germany}
}
\renewcommand{\shortauthors}{Sch{\"o}ne et al.}

\begin{abstract}
  Event-based sensors are well suited for real-time processing due to their fast response times and encoding of the sensory data as successive temporal differences. 
    These and other valuable properties, such as a high dynamic range, are suppressed when the data is converted to a frame-based format.
    However, most current methods either collapse events into frames or cannot scale up when processing the event data directly event-by-event.
    In this work, we address the key challenges of scaling up event-by-event modeling of the long event streams emitted by such sensors, which is a particularly relevant problem for neuromorphic computing.
    While prior methods can process up to a few thousand time steps,  our model, based on modern recurrent deep state-space models, scales to event streams of millions of events for both training and inference.
    We leverage their stable parameterization for learning long-range dependencies, parallelizability along the sequence dimension, and their ability to integrate asynchronous events effectively to scale them up to long event streams.
    We further augment these with novel event-centric techniques enabling our model to match or beat the state-of-the-art performance on several event stream benchmarks.
    In the Spiking Speech Commands task, we improve state-of-the-art by a large margin of \SI{7.7}{\percent} to \SI{88.4}{\percent}.
    On the DVS128-Gestures dataset, we achieve competitive results without using frames or convolutional neural networks.
    Our work demonstrates, for the first time, that it is possible to use fully event-based processing with purely recurrent networks to achieve state-of-the-art task performance in several event-based benchmarks.
\end{abstract}

\ccsdesc[500]{Computing methodologies~Machine learning algorithms}

\keywords{Event-stream modeling, state-space models, event-based vision, deep learning, neuromorphic sensors}

\maketitle

\section{Introduction}
Inspired by the sensory systems in biology, neuromorphic sensors implement an asynchronous and event-based encoding of local environmental changes \citep{Lichtsteiner2008, Posch2011, Chan2007, Caviglia2017, PropheseeGen4}.
This sensing paradigm promises several advantages over classical sensors, including energy efficiency, low latency, increased temporal resolution, and dynamic range.
For example, systems subject to rapidly changing environments or lighting conditions, such as autonomous robots, benefit from the high dynamic range and low latency of event-based vision sensors.
Subsequent processing stages such as machine learning systems, must be compatible with the sensors' asynchronous and temporally sparse event-streams to fully leverage the neuromorphic sensing paradigm.

However, machine learning methods struggle to effectively handle event-streams asynchronously in an event-by-event processing setting. 
This is due to three key challenges of working with neuromorphic event-streams:
(1) Integrating neuromorphic signals event-by-event requires learning interactions between events far apart in time and/or spatial dimensions.
This is the well-known problem of learning \emph{long-range dependencies} that has been extensively studied in the recurrent neural networks literature \citep{Hochreiter1991, Bengio1994}.
(2) Neuromorphic sensors can emit large numbers of events per second, generated in parallel from up to a million asynchronous input channels \citep{PropheseeGen4}. 
Effectively processing very long sequences from neuromorphic sensors requires \emph{parallelization} to use sparsity and asynchrony effectively. 
The vast advances of highly parallel hardware accelerators favor sequence modeling methods that allow parallelization along the sequence length.
(3) \emph{Asynchronous processing.} Neuromorphic sensors produce events in irregular time intervals from many asynchronous sensor channels.
Most modern machine learning algorithms require a fixed step size to process sequences effectively.
Continuous-time methods have been developed to handle irregular sequences \citep{Schirmer2022}, but struggle with very long sequences due to a limited ability to learn long-range dependencies.
Ultimately, machine learning systems that use event-based sensors today often collapse events into frames and thus lose many of the advantages of direct event-based processing.

This work demonstrates the first scalable machine learning method to effectively learn event-based representations directly from high-dimensional asynchronous event-streams. 
Our method uses linear state-space models (SSMs), a class of machine learning models that have successfully modeled complex sequential data \citep{Gu2022S4}. 
They are a type of recurrent neural network that can be efficiently parallelized along the sequence dimension (challenge 2).
Together with their ability to model long-range dependencies (challenge 1), this property allows significant improvements on tasks like sequential image processing \citep{Gu2022S4, Smith2023} and raw-audio processing \citep{Goel2022}. 
However, asynchronous integration of inputs (challenge 3) has not been addressed by the literature.
Furthermore, the machine learning tasks covered by the literature require modeling sequences of a few thousand up to a hundred thousand steps, while neuromorphic sensory signals pose examples of even longer sequences ranging up to millions of events.
This work addresses modeling of long asynchronous time-series (challenge 3), while maintaining long-range dependency learning and parallelization as introduced by the SSM literature.
We propose several novel techniques on top of SSMs and apply them directly to process neuromorphic sensor signals event by event.
We demonstrate that this effectively mitigates all three challenges of (1) long-range dependencies, (2) parallelization, and (3) asynchronous processing. 
Remarkably, the state-space model extracts spatio-temporal features from event-based vision streams without any convolutional layers.

To our knowledge, this is the first scalable event-by-event processing method for neuromorphic event-streams that achieves compelling results compared to previous frame-based approaches.

\section{Related Work}
\label{sec:related-work}
Learning representations from neuromorphic event-streams requires methods that handle very long sequences and sequences with events that are irregularly sampled in time from a set of asynchronous sources.
Existing methods represent event-streams as time-frames and learn these representations either end-to-end \citep{Gehrig2019}, or construct them manually \citep{Lagorce2017, Innocenti2021, Barchid2022, Liu2022}.
The time-frame representation allows the processing of the data with convolutional neural networks (CNNs) as well as recurrent neural networks (RNNs) and their spiking variants (SNNs). 
Given a particular time-frame representation, events can be integrated into the frame representation asynchronously \citep{messikommer2020, Cordone_2021_IJCNN}.
\citet{Zubic2024} apply state-space models to frames extracted from an event-based vision sensor to speed up training.
An exception from the frame-based paradigm is \citet{martinturrero2024}, who collect a set of events into a learned tensor representation that can be updated asynchronously to allow asynchronous inference on event-streams.
However, we significantly outperform their asynchronous method on DVS128-Gestures and even outperform their synchronous method, while using much fewer parameters.
To the best of our knowledge, ours is the first work that operates fully asynchronously on the event-stream and at the same time scales to state-of-the-art performance on standard neuromorphic benchmarks.
\minisec{Spiking neural networks}
Spiking neural networks are often formulated as continuous-time models and discretized for simulation, much like state-space models.
The continuous-time formulation theoretically allows asynchronous event-based simulations.
In practice, however, most researchers discretize on relatively coarse-grained equidistant time grids.
For example, the spiking audio models of \citet{Bittar2022} and \citet{hammouamri2024learning} use simulation steps of $\Delta t = \SI{25}{\milli\second}$ and $\Delta t = \SI{10}{\milli\second}$ respectively.
Vision models even use simulation steps of up to $\Delta t = \SI{100}{\milli\second}$ \citep{Liu2022} or just simulate a total of 4 steps on a single sample \citep{Fang2023}.
In contrast, our method integrates every single event asynchronously with arbitrary resolution without sacrificing simulation scalability.
\minisec{Long sequences}
The problem of modeling long sequences has been addressed by recent developments in linear state-space models (SSMs).
SSM-like models were first proposed and shown to have long-range memory in \citet{Voelker2019LMU} and have since been developed to be highly effective scalable models~\citep{Gu2022S4} via structured linear transformations that efficiently parallelize on modern accelerators.
SSMs have successfully been scaled to very long sequences such as raw-audio processing of up to \num{128000} steps \citep{Goel2022}, and for autoregressive DNA modeling consisting of over a over a million steps using a time-variant SSM \citep{Gu2023}.
Time-variant diagonal SSMs have recently demonstrated exceptional performance on large-scale language modeling \citep{Gu2023, De2024}.
The parallelization property of linear SSMs has since been applied to parallelize training of neuromorphic systems that operate on time-frames \citep{Yarga2023, Fang2023}.
\minisec{Irregular sequences}
Conventional deep learning models are unsuitable for long, irregular time series.
They either lack effective time-coding paradigms, such as CNNs or RNNs, or suffer from unfavorable time complexity, such as self-attention-based models.
An early attempt to learn long-range dependencies in irregular time series, such as neuromorphic event-streams, was presented by \citet{Neil2016}.
They added a new gating mechanism to recurrent architectures that allowed the model to attend to specific frequencies in the data.
\citet{Schirmer2022, Ansari2023} leveraged continuous-time state-space models driven by stochastic differential equations that integrate discrete observations in a probabilistic manner.
\citet{Smith2023} show that deterministic state-space models can solve the single source pendulum toy task,  which was already used by \citet{Schirmer2022} while maintaining the favorable properties of SSMs discussed above.
Neuromorphic event-streams are at an entirely different scale than the benchmarks used in deep learning papers.
They feature up to a million asynchronous source channels and can sample millions of events per second \citep{Perot2020}.
We refine the work of \citet{Smith2023} to handle asynchronous irregular time series of this scale and show promising results on neuromorphic benchmark datasets.

\section{Scalable Event-stream Modeling with Deep State-Space Models}

\begin{figure*}[t]
    \centering
    \includegraphics[width=\textwidth]{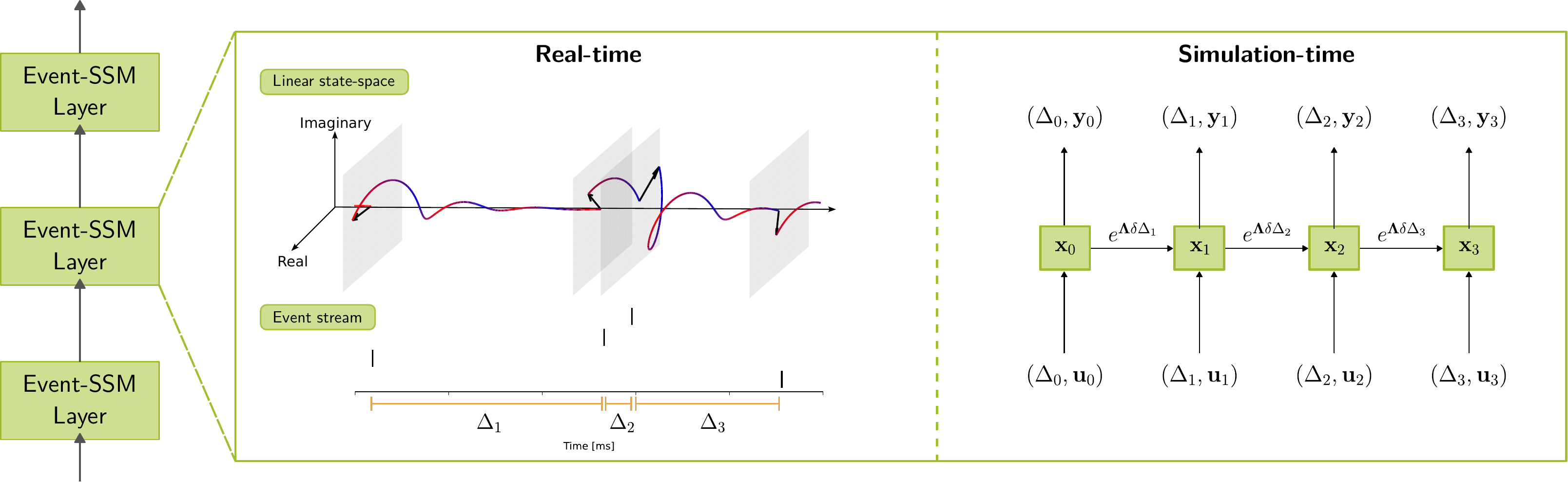}
    \caption{
        In real-time, our model evolves a linear time-invariant state-space in continuous time and integrates the delta-coded event-stream along the way.
        The strength of our model stems from its duality with a linear time-variant recurrence relation discretized over the event times.
        This allows the simulation to leverage the associative scan primitive to parallelize the dynamical system over time.
    }
    \label{fig:figure1}
\end{figure*}
We focus on the \emph{simplified state-space layer} (S5) \citep{Smith2023} due to its favorable trade-off between simplicity and efficiency for the set of tasks of interest to us. 
In \secref{state-space-models}, we review the S5 model, and in \secref{event-stream-modeling}, we show how it can be used for efficient, scalable event-stream modeling. In \secref{related-work} we briefly review other related state-space model architectures. 
\subsection{Deep State-Space Models}
\label{sec:state-space-models}
A linear time-invariant state-space model (SSM) in continuous time is given by the linear system
\begin{align}
    \bfxdot (t)
    &= \bfA \bfx(t) + \bfB \bfu(t) \label{eq:ssm-x} \\
    \bfy(t) 
    &= \bfC \bfx(t) + \bfD \bfu(t) \label{eq:ssm-y}\,,
\end{align}
where ${\bfx\in\R{H}}$ is the state-space vector and $\bfxdot$ is its time derivative, $\bfy\in\R{N}$ is the output vector, ${\bfu\in\R{N}}$ is the input signal, and ${\bfA\in\R{H\times H}}$, ${\bfB\in\R{H\times N}}$, ${\bfC\in\R{N\times H}}$, ${\bfD\in\R{N\times N}}$ are the learnable parameters of the system.
Effectively modeling event-streams requires learning dependencies between distant events and irregularly sampled events.
Both requirements can be addressed with the above continuous-time model.
The extensive theory on linear time-invariant systems allows us to reason efficiently about the system's long-term stability.
The continuous-time dynamics can also be discretized on any set of timestamps for irregular sequence modeling.
We will elaborate on both features below.
\minisec{Long-range dependencies}
Recurrent sequence models, such as recurrent neural networks, fundamentally suffer from a trade-off between long-term memory and vanishing gradients \citep{Hochreiter1991, Bengio1994}.
SSMs mitigate this problem by avoiding non-linear recurrence. 
In contrast to most non-linear dynamics, the linear system in eq.~\eqref{eq:ssm-x} can be carefully tuned for effective long-range dependency modeling.

The set of real matrices that are \emph{diagonalizable} over the complex numbers are dense in the space of real matrices,
i.e. a randomly initialized state-space model is diagonalizable almost certainly over the complex numbers.
In this case, there exists a diagonal matrix $\bfLambda\in\C{H\times H}$ and an invertible projection $\bfP\in\C{H\times H}$, such that $\bfA = \bfP\bfLambda \bfP^{-1}$.
Hence, there is an equivalent diagonalized SSM with state space variable $\bftilde{x}$ such that $\bfx(t) = \bfP^{-1}\bftilde{x}(t)$.
We summarize that the class of state-space models defined by
\begin{align}
    \bfxdot (t)
    &= \bfLambda \bfx(t) + \bfB \bfu(t) \label{eq:ssm-x-diagonal} \\
    \bfy(t)
    &= \mathfrak{Re}\left( \bfC \bfx\left(t\right) \right) + \bfD \bfu(t) \label{eq:ssm-y-diagonal} \,
\end{align}
covers the same dynamics as eqs.~\eqref{eq:ssm-x} and \eqref{eq:ssm-y},
where $\mathfrak{Re}(\bfz)$ denotes the real part of a complex variable $\bfz$, ${\bfLambda\in\C{H\times H}}$ is diagonal and ${\bfB\in\C{H\times N}}$, ${\bfC\in\C{N\times H}}$, ${\bfD\in\R{N\times N}}$.

The diagonalized system comes with computational and conceptual advantages.
While applying the recurrent operator $\bfA$ requires \bigorder{H^2} operations per step, the diagonal operator $\bfLambda$ requires only \bigorder{H} operations per step.
Furthermore, the $H$ learnable parameters of the state-to-state operator $\bfLambda$ are precisely its spectrum.
Since the spectrum defines the system's long-term behavior, we can effectively control its long-term behavior by carefully parameterizing $\bfLambda$.

\begin{figure*}
    \centering
    \includegraphics[width=\textwidth]{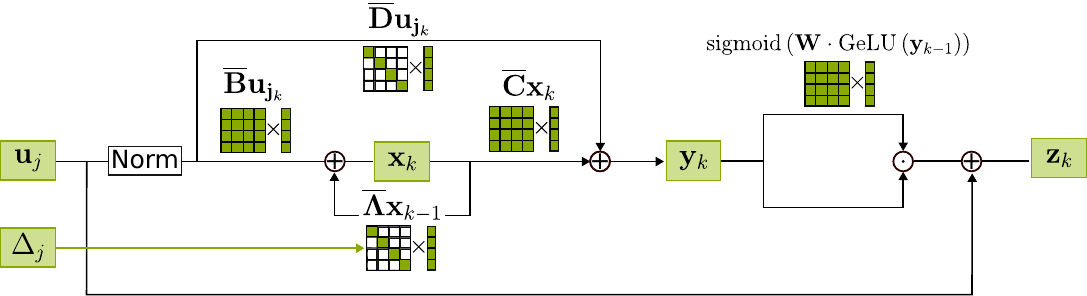}
    \caption{
    Our modified simplified state-space layer consists of an SSM followed by a non-linear multiplicative transformation.
    A skip connection and a normalization layer complete the block.
    The information about event timings is passed to the model via the differences $\Delta_i=t_i-t_{i-1}$.
    }
    \label{fig:S5}
\end{figure*}

Long-range dependencies in $\bfx(t)$ can be modeled effectively if the entries of $\bfLambda$ have negative real part, i.e. they reside on the left half-plane of $\C{}$.
\citet{Gu2022Param, Smith2023, Orvieto2023a} enforce the left-half plane condition on the spectrum by parameterizing $\bfLambda = -\exp(\bfPhi) + i \bfTheta$, with potentially diagonal matrices $\bfPhi,\bfTheta\in\R{H\times H}$.
While any positive activation function could force the real part of the eigenvalues to be negative, the exponential function is used throughout the literature.
\minisec{Discretization}
Most prior works discretize the continuous system in eqs.~\eqref{eq:ssm-x-diagonal} and \eqref{eq:ssm-y-diagonal} on a regular grid $t_0,\dots, t_M$ with ${t_m - t_{m-1} \equiv \Delta > 0}$.
\citet{Smith2023} discretize S5 with the zero-order hold method, which yields a discrete-time system
\begin{align}
    \bfx_k &= \bfbar{\Lambda} \bfx_{k-1} + \bfbar{B} \bfu_k \label{eq:S5-x} \\
    \bfy_k &= \mathfrak{Re}\left(\bfbar{C} \bfx_k\right) + \bfbar{D} \bfu_k \label{eq:S5-y} \,,
\end{align}
where $\bfx_k = \bfx(t_k), \bfu_k = \bfu(t_k)$, $\Delta$ is the step size and
\begin{align}
    \bfbar{\Lambda} = e^{\bfLambda \Delta}\,, \quad \bfbar{B} = \bfLambda^{-1}\left(\bfbar{\Lambda} - 1 \right) \bfB\,, \quad \bfbar{C} = \bfC\,, \quad \bfbar{D} = \bfD \,. \label{eq:zoh-param}
\end{align}
This discretization method yields two properties that were later found essential for long-range modeling by~\citet{Orvieto2023a}.
Firstly, the state-to-state operator $\bfbar{\Lambda}$ is parameterized by an exponential, which enhances stability in conjunction with the half-plane parameterization described above.
Secondly, the input to the state-space $\bfbar{B}\bfu_k$ is modulated by a coefficient that depends on the spectrum of the $\bfLambda$.
This coefficient balances the magnitudes of the different components of $\bfx_k$ taking their respective effective time scales into account,
which leads to more stable learning dynamics on very long sequences.
We elaborate an extension for asynchronous sensors in \secref{event-stream-modeling}.
\minisec{Simplified state-space layers}
The S5 model \citep{Smith2023} is a stack of simplified state-space layers, depicted in \figref{S5}.
The S5 layer consists of the linear state-space model as described in eqs.~\eqref{eq:S5-x} and \eqref{eq:S5-y} and a non-linear multiplicative transformation.
According to eq.~\eqref{eq:zoh-param}, ${\bfbar{\Lambda} = e^{\left(-e^{\bfPhi} + i\bfTheta\right)\Delta}}$ and ${\bfbar{B} = \bfLambda^{-1}\left(\bfbar{\Lambda} - 1 \right) \bfB}$.
The learnable parameters are ${\bfPhi,\bfTheta\in\R{H\times H}}$, ${\bfB\in\C{H\times N}}$, ${\bfC\in\C{N\times H}}$, ${\bfD\in\R{N\times N}}$, where $\bfPhi,\bfTheta,D$ are diagonal matrices.
A non-linear multiplicative interaction similar to the GLU activation \citep{GLU2017} is applied to the SSM output
\begin{align}
    \bfv_k &= \operatorname{GeLU}\left(\bfy_k\right)\\
    \bfz_k &= \bfy_k \odot \operatorname{sigmoid}\left(\bfW\bfv_k\right) \,.
\end{align}
In addition, skip connections and normalization layers are used in S5, which is in line with most modern deep learning models.

\minisec{Parallelization}
Parallelization is a critical component of modern deep learning systems.
Non-linear recurrent neural networks such as LSTMs or biologically plausible spiking neural networks lack parallelization along the sequence lengths.
Such models have significantly restricted throughput on highly parallel processors such as GPUs.
Therefore, training them has been limited to small models or datasets,  posing a major drawback in the modern era of scalable deep learning.

Whereas the linear recurrence of eq.~\eqref{eq:S5-x} allows for efficient parallelization along the sequence in addition to training stability.
Therefore, variants of linear state-space models such as S5 learn long-range dependencies in sequences, and scale computationally to very long sequences up to a hundred thousand time steps \citep{Goel2022}.
A comprehensive treatment of the parallelization of linear recurrence equations based on prefix sums of associative operators can be found in \citet{blelloch1990prefix}.
Consider the first-order recurrence relation similar to eq.~\eqref{eq:S5-x}
\begin{align}
    \bfx_k = \begin{cases}
        \bfbar{B}\bfu_0 & i = 0 \\
        \bfbar{\Lambda} \cdot \bfx_{k-1} + \bfbar{B}\bfu_k & 0 < i \leq T
    \end{cases} \,. \label{eq:first-order-recurrence}
\end{align}
Let $\bfc_k = \left(\bfa_k, \bfb_k\right)$. 
As shown in \citet{blelloch1990prefix}, the operation in eq.~\eqref{eq:first-order-recurrence} can be reduced to an associative operator
\begin{align}
    \bfc_k \otimes \bfc_l = \left(\bfa_k\bfa_l, ~ \bfa_k \bfb_l + \bfb_k\right) \,.
\end{align}
Hence, the associative scan primitive resolves a recurrence of length $M$ in \bigorder{\log M} time given sufficiently many parallel processors.

\subsection{Scalable Event-stream Modeling}
\label{sec:event-stream-modeling}
In this subsection, we show that these advantageous properties of S5 are also useful in the case of asynchronous event-streams.
An event-stream is an ordered set $E=\left\{(t_m, j_m) ~\vert~ m=0,\dots,M\right\}$ of event times $t_0<\dots<t_M\in\R{}$ and corresponding event source channels $j_m\in\{1,\dots,J\}$.
As we will see in the following, our event-based SSM operates efficiently on the differences $\Delta_m = t_m - t_{m-1}$ instead of the timestamps themselves.
A linear projection translates the integer representation of event sources $j_m$ to the model's vector representation via ${\bfu_m = \bfE \cdot\operatorname{onehot}(j_m)}$ for $m\in\{0, \dots, M\}$.
This operation can be efficiently implemented as a look-up table that queries the $j_m$-th column from the projection matrix $\bfE\in\R{J\times N}$, 
a common practice in language modeling.
\minisec{Discretization on irregular event-streams}
Consider a set of $J$ asynchronous channels.
In continuous time, an event-stream can then be represented as a sum of dirac deltas
\begin{align}
    \bfu(t)
    = \sum_{m=0}^M \delta\left(t - t_m\right)\bfu_m \label{eq:dirac-coding}\,.
\end{align}
The general analytical solution of the ODE in eq.~\eqref{eq:ssm-x-diagonal} with initial conditions $\bfx\left(t_0\right) = 0$ for the delta coded input as in eq.~\eqref{eq:dirac-coding} is
\begin{align}
    \bfx(t_k) = \integrate{e^{\bfLambda(t_k-s)}\bfB \bfu(s)}{s}{t_0}{t_k}  =  \sum_{m=0}^M e^{\bfLambda(t_k-t_m)}\bfB \bfu_m
    \label{eq:analytic-integral} \,.
\end{align}
This solution admits a recursive formulation
\begin{align}
    \bfx_k = \bfx(t_k) 
    &= \sum_{m=0}^M e^{\bfLambda(t_k-t_m)}\bfB \bfu_m \nonumber \\
    &= e^{\bfLambda(t_k - t_{k-1})} \left( \sum_{m=0}^{M-1} e^{\bfLambda(t_{k-1}-t_m)}\bfB \bfu_m \right) + \bfB \bfu_k \nonumber\\
    &= e^{\bfLambda\Delta_k}\bfx_{k-1} + \bfB \bfu_k \label{eq:discrete-ssm} \,,
\end{align}
where $\Delta_k = t_k - t_{k-1}$.
We obtain a formalism to process the irregularly sampled event-stream with a discrete (linear) recurrent neural network, whose state update depends on both the event times via $\Delta_k$ and the input values $\bfu_k$.
Notably, this RNN can be simulated fully event-based in the discrete time domain for both inference and learning.
Since the RNN state is just a linear combination of the input events, 
there is no need for advanced event-timing dependent gradient computation methods such as EventProp \citep{wunderlich2021event}.
\minisec{Input normalization of asynchronous events}
In \secref{state-space-models}, we discuss the benefits of the input normalization factor $\bfLambda^{-1}\left(\bfbar{\Lambda} - 1\right)$ that emerges from the zero-order hold discretization for stable learning.
Since the normalization factor depends on the time step $\Delta$ through $\bfbar{\Lambda}$, the inputs are effectively weighted w.r.t. their timing relative to other inputs.
In contrast, we argue that asynchronous events should be independently integrated into the state-space.
The normalization factor of an event should not depend on the relative timings of other asynchronous events.
Therefore, we disentangle the $\Delta$ in \eqref{eq:zoh-param} as $\Delta = \bfdelta \Delta_k$, where $\bfdelta$ is a new learnable parameter and $\Delta_k = t_k - t_{k-1}$ are the actual differences of time steps.
Compared to eq.~\eqref{eq:zoh-param}, we obtain our asynchronous discretization method with an event time independent normalization factor
\begin{align}
    \bfx_k &= \bfbar{\Lambda}_k \bfx_{k-1} + \bfbar{B} \bfu_k \label{eq:event-S5-x} \\
    \bfy_k &= \mathfrak{Re}\left(\bfbar{C} \bfx_k\right) + \bfbar{D} \bfu_k \label{eq:event-S5-y} \,,
\end{align}
with
\begin{align}
    \bfbar{\Lambda}_k = e^{\bfLambda \bfdelta \Delta_k} \,, \quad \bfbar{B} = \bfLambda^{-1} \left(e^{\bfLambda \bfdelta} - 1 \right) \bfB \,, \quad \bfbar{C} = \bfC \,, \quad \bfbar{D} = \bfD\,.
    \label{eq:async-discretization}
\end{align}
The inputs are therefore normalized by the units' individual time scales $\tau = \frac{1}{\delta}$, but not by the event-timing dependent $\Delta_k$.
We provide evidence supporting this strategy in \tabref{ablations}

Note that the trainable parameters are $\bfdelta\in\R{H\times H}_{+},\bfPhi,\bfTheta\in\R{H\times H},\bfB\in\C{H\times N}, \bfC\in\C{N\times H}, \bfD\in\R{N\times N}$, where again $\bf\Phi,\bfTheta,\bfD$ and $\bfdelta$ are diagonal matrices.
\minisec{Parallelization}
With the parameterization in eq.~\eqref{eq:async-discretization}, the system described by eq.~\eqref{eq:S5-x} and eq.~\eqref{eq:S5-y} becomes a linear time-variant system.
The associative scan is still a valid parallelization primitive, since 
\begin{align}
    \bfbar{\Lambda}\left(\Delta_k\right) \cdot \bfbar{\Lambda}\left(\Delta_l\right) 
    &= e^{\bfLambda \bfdelta \Delta_k} e^{\bfLambda \bfdelta \Delta_l} \nonumber \\
    &= e^{\bfLambda \bfdelta (\Delta_k + \Delta_l)} = \bfbar{\Lambda}\left(\Delta_k + \Delta_l\right) \,.
\end{align}
Therefore, the operator
\begin{align}
    \bfc_k \otimes \bfc_l 
    &=  \left(\bfbar{\Lambda}(\Delta_k), \bfb_k\right) \otimes \left(\bfbar{\Lambda}(\Delta_l), \bfb_l \right) \nonumber \\
    &= \left( \bfbar{\Lambda}(\Delta_k + \Delta_l),~\bfbar{\Lambda}(\Delta_k) \bfb_l + \bfb_k \right)
\end{align}
acting on $\bfc_k = \left(\bfbar{\Lambda}(\Delta_k), \bfb_k\right)$ is associative as well, 
and can be used to parallelize the recurrent system described by eqs.~\eqref{eq:event-S5-x} and \eqref{eq:event-S5-y} with parameterization as in eq.~\eqref{eq:async-discretization}.
\minisec{Event-pooling architecture}
Event-by-event processing becomes expensive when the model size and the sequence length are scaled up.
Although the recurrent operations are cheap in the case of diagonal SSMs, every processed event propagates through the network's numerous dense feed-forward transformations.
Furthermore, saving the activations of every event for backpropagation through time causes accelerators to run out of memory when training on long sequences.
We mitigate this issue by introducing an event-pooling mechanism that, by subsampling the sequence, can drastically reduce the required compute and memory.
Subsampling architectures are widely used in vision models and were proposed for recurrent architectures in \citet{Graves2008}.
They have also been used in state-space models (e.g. \citep{Goel2022}).
A sequence of lengths $M$ with vectors of dimension $H$ is compressed to a sequence of length $M / p$.
Oftentimes, the vector dimension is increased upon sequence subsampling to $H q$.
Since linear recurrences effectively compress information \citep{Orvieto2023}, 
we decided to apply event-pooling after each state-space layer.
Hence, $M$ inputs $\bfu_m$ are integrated into the state-space $\bfx$, but only a subsampled sequence of length $M / p$ is forwarded to the linear transformation $\bfbar{C}\bfx$.

Similar to frame-based methods, our subsampling architecture reduces the computational overhead by pooling a set of events.
In the context of continous-time state-space models, subsampling is equivalent to averaging over the spatio-temporal representation computed by the state-space eq.~\eqref{eq:event-S5-x} for $p$ consecutive events.
While converting events into frames is a preprocessing step, subsampling can be applied in multiple layers of the model to form hierarchical representations as common practice in audio and vision models. 

\section{Experiments}
We evaluate our method, \textbf{Event-SSM}, on three event-based datasets that are popularly used in the neuromorphic community.
The datasets are provided as raw event-streams, which we process directly without preprocessing into frames.
The Spiking Heidelberg Digits (SHD) and Spiking Speech Commands (SSC) datasets were proposed to standardize the evaluation of neuromorphic models \citep{Cramer2022}, both consisting of spike trains that were converted from microphone recordings.
DVS128 Gestures (DVS) is a small-scale action recognition dataset \citep{Amir2017} consisting of a set of 11 gestures recorded with a dynamic vision sensor in $128\times 128$ pixels resolution.
While the number of samples in SSC exceeds the other two datasets by an order of magnitude, the number of events per sample in the DVS dataset exceeds the two audio datasets by more than an order of magnitude.
An overview of the statistics of the three datasets is presented in \tabref{datasets}.

All models presented in this work feature six simplified state-space layers as depicted in \figref{S5},
with state sizes of either $H=64$ or $H=128$.
To improve generalization, we implemented an event-based variant of CutMix data augmentation \citep{CutMix}.
Additional samples were generated by randomly mixing existing ones, i.e. a contiguous stream of events was randomly mixed into another sample of the same batch.
Labels were mixed according to the relative number of events from both samples.
We implement our model in JAX \citep{jax2018github}.
The efficient parallelism of our method allows us to train on the larger SSC (\SI{600}{\million} events per epoch) and DVS (\SI{390}{\million} events per epoch) datasets in 2 – 10 h on a single A100 GPU.
For implementation details and precise hyperparameters, we refer the reader to our published code repository~\footnote{\url{https://github.com/Efficient-Scalable-Machine-Learning/event-ssm}}.

By reviewing the code of published papers included in our baselines in \tabref{shd} and \tabref{dvs-gestures},
we found that it is common practice to pick the best model based on the test set instead of a separate validation set.
We did the same for a fair comparison with the baseline methods, even though we believe that this procedure does not align with best practice.
\begin{table}
    \centering
    \begin{tabular}{l S[table-format=2.0] S[table-format=6.0] S[table-format=6.0]}
    \toprule
         \multirow{2}{*}{Dataset} & {\multirow{2}{*}{Classes}} & {Training}  & {Median number}  \\
         & & {samples} & {of events} \\
    \midrule
        Spiking Heidelberg Digits & 20 & 8200 & 8000 \\ 
        Spiking Speech Commands & 35 & 75500 & 8100 \\
        DVS128 Gestures    & 11 & 1100 & 300000 \\
    \bottomrule
    \end{tabular}
    \caption{The datasets used to evaluate our event-stream modeling method differ in the number of samples present in the dataset as well as the number of events per sample per dataset.
    All values are given to two significant digits.
    }
    \label{tab:datasets}
\end{table}
\begin{figure}
    \centering
    \includegraphics[width=\linewidth]{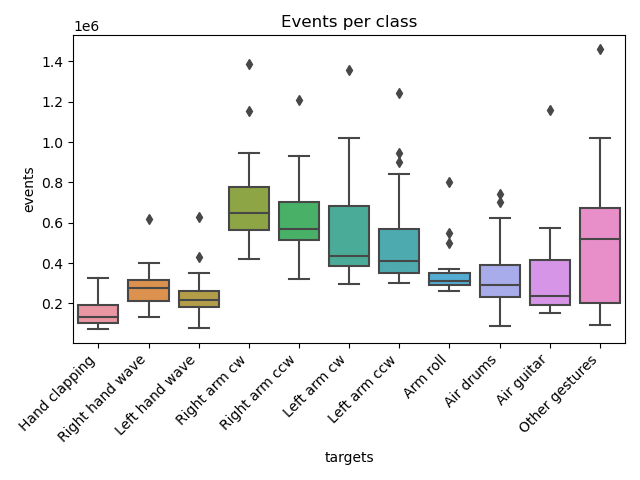}
    \caption{
        Distribution of the number of events per class in the DVS128-Gesture dataset.
        The median number of events per sample is about 300,000, 
        and the maximum number of events per sample is about 1.5 million. 
    }
    \label{fig:dvs-data}
\end{figure}
\subsection{Spiking audio processing}
We present a new state-of-the-art classification result on both spiking audio datasets.
These results show that the exact integration of spike timings can, in fact, improve the performance of spiking audio models.
A combination of time-jitter, channel-jitter, random noise, drop-event, and cut-mix data augmentations was applied to improve generalization.
\Tabref{shd} shows our results on the SHD dataset.
We note that model performance is almost saturated on this dataset.
Furthermore, we observed a larger variance with different random seeds compared to the larger SSC dataset.

This leads us to the conclusion that the much less saturated and larger scale SSC dataset is more appropriate for evaluating state-of-the-art methods.

Results for SSC are shown in \tabref{ssc}.
Our method significantly outperforms the state-of-the-art set recently by \citet{hammouamri2024learning} by a margin of almost \SI{6.6}{\percent}, while using much fewer parameters.
\begin{table}[]
    \centering
    \begin{tabular}{l c S[table-format=2.1] S[table-format=2.1]}
        \toprule
        \textbf{Spiking Heidelberg} & Async. & {Test} & {Num} \\
        \textbf{Digits} & events & {accuracy} & {params} \\
        \midrule
        \citet{Bittar2022} & \ding{55} & \SI{93.1}{\percent} & \SI{0.1}{\million}\\
        \citet{Bittar2022} & \ding{55} & \SI{94.6}{\percent} & \SI{3.9}{\million}\\
        \citet{hammouamri2024learning} & \ding{55} & \SI{95.1}{\percent} & \SI{0.2}{\million} \\
        \midrule
        \textbf{Event-SSM} & \ding{51} & \textbf{\SI{95.9}{\percent}} & \SI{0.4}{\million} \\
        \bottomrule
    \end{tabular}
    \caption{
        Comparison of our Event-SSM to the state-of-the-art on the Spiking Heidelberg Digits dataset~\citep{Cramer2022}.
    }
    \label{tab:shd}
\end{table}
\begin{table}[]
    \centering
    \begin{tabular}{l c S[table-format=2.1] S[table-format=2.1]}
        \toprule
        \textbf{Spiking Speech} & Async. & {Test} & {Num} \\
        \textbf{Commands} & events & {accuracy} & {params} \\
        \midrule
        \citet{Bittar2022} & \ding{55} & \SI{71.7}{\percent} & \SI{0.1}{\million}\\
        \citet{Bittar2022} & \ding{55} & \SI{77.4}{\percent} & \SI{3.9}{\million}\\
        \citet{hammouamri2024learning} & \ding{55} & \SI{79.8}{\percent} & \SI{0.7}{\million} \\
        \citet{hammouamri2024learning} & \ding{55} & \SI{80.7}{\percent} & \SI{2.5}{\million} \\
        \midrule
        \textbf{Event-SSM} & \ding{51} & \SI{85.3}{\percent} & \SI{0.1}{\million} \\
        \textbf{Event-SSM} & \ding{51} & \textbf{\SI{88.4}{\percent}} & \SI{0.6}{\million} \\
        \bottomrule
    \end{tabular}
    \caption{
        Comparison of our Event-SSM to the state-of-the-art on the Spiking Speech Commands dataset \citep{Cramer2022}.
        }
    \label{tab:ssc}
\end{table}

\subsection{Event-based vision processing}
The DVS128 Gestures dataset was recorded with a dynamic vision sensor of $128\times 128$ resolution \citep{Lichtsteiner2008}.
Each pixel is represented by two channels, encoding the two event polarities, resulting in $C=128\times 128 \times 2 = 32768$ asynchronous channels for the DVS128 Gestures dataset.
The large number of asynchronous channels results in a very large number of events per second.
An overview of the distribution of the number of events per sample across the classes of the DVS dataset can be obtained from \figref{dvs-data}.
Consequently, learning representations for event-based vision is one of the largest scale benchmarks for event-based processing systems.
The previous best-performing baseline models collected events into a 4-d tensor representation which was then processed by convolutional neural networks composed of artificial or spiking neurons.
Doing this mitigates the computational overhead of processing every event individually, and circumvents the need to process very long sequences of irregularly sampled events.
In contrast,  our event-based state-space model directly processes the event-stream recorded from the dynamic vision sensor with a recurrent neural network, without using spatial convolutions.
Spatio-temporal representations are solely learned from the linear state-space model and non-linear feedforward transformations.
Despite breaking with the pervasive convention of binning events into time-frames and processing with CNNs, our event-based state-space model achieved competitive results as reported in \tabref{dvs-gestures}.
To improve generalization, a combination of data augmentation methods such as spatial-jitter, time-jitter, random noise, drop-event, geometric augmentations \citep{Li2022}, and cut-mix were applied.

Training recurrent networks with backpropagation through time (BPTT) requires storing (or recomputing) the activations for every step in the sequence.
The large number of events per sample, therefore, quickly saturated GPU memory.
To fit reasonable batch sizes into the \SI{40}{\giga\byte} HBM memory of our A100s, 
we sliced the training data into shorter sequences.
Yet, evaluation was conducted on full samples of up to \SI{1.5}{\million} events.
Surprisingly, we found that training on slices of \num{32768} events suffices to reach the baseline performance.
\begin{table}[]
    \centering
    \begin{tabular}{l c S[table-format=2.1] S[table-format=2.1]}
        \toprule
        \textbf{DVS128} & Async. & {Test} & {Num} \\
         \textbf{Gesture} & events & {accuracy} & {params} \\
        \midrule
        \citet{Yousefzadeh2019} & \ding{55}  & \SI{95.2}{\percent} & \SI{1.2}{\million} \\
        \citet{Xiao2022} & \ding{55} &\SI{96.9}{\percent} & {-} \\
        \citet{subramoney2023efficient} & \ding{55} & \SI{97.8}{\percent} & \SI{4.8}{\million} \\
        \citet{She2022} & \ding{55} & \SI{98.0}{\percent} & \SI{1.1}{\million} \\
        \citet{Liu2022} & \ding{55} & \textbf{\SI{98.8}{\percent}} & {-} \\
        \citet{martinturrero2024} & \ding{55} & \SI{96.2}{\percent} & \SI{14}{\million} \\
        \citet{martinturrero2024} & \ding{51} & \SI{94.1}{\percent} & \SI{14}{\million} \\
        \midrule
        CNN~+~S5 (time-frames) & \ding{55} & \SI{97.8}{\percent} & \SI{6.8}{\million} \\
        CNN~+~S5 (event-frames) & \ding{55} & \SI{97.3}{\percent} & \SI{6.8}{\million} \\
        \midrule
        \textbf{Event-SSM} & \ding{51} & \SI{97.7}{\percent} & {\SI{0.8}{\million} + \SI{4.2}{\million}} \\
        \bottomrule
    \end{tabular}
    \caption{
        Comparison of our Event-SSM to the state-of-the-art on the DVS128-Gesture dataset \citep{Amir2017}.
        We report our model's number of parameters as the parameters of the SSM + embedding look-up.
        Due to the sensor resolution, most parameters are learned embedding vectors rather than SSM parameters.
        }
    \label{tab:dvs-gestures}
\end{table}
\subsection{Ablation study}
\label{sec:ablation-study}
In \secref{event-stream-modeling}, we argued that naively applying the discretization methods of most SSM works to asynchronous event-streams is not ideal.
\Tabref{ablations} compares our method as presented in eq.~\eqref{eq:async-discretization} with the popular zero-order hold (ZOH) method employed by \citet{Smith2023}, the naive integration of Dirac delta pulses in eq.~\eqref{eq:dirac-coding}, and vanilla S5 without passing in the timestamps at all.
Integrating events according to eq.~\eqref{eq:async-discretization} clearly improves the performance over the other methods.
These results provide further evidence that event timings can improve representations of event-based systems.
\begin{table}[]
    \centering
    \begin{tabular}{l S[table-format=2.1]}
    \toprule
         Model & Accuracy \\
    \midrule
         \textbf{Event-SSM} & \SI{86.9+-0.4}{\percent} \\
         S5 with Dirac discretization & \SI{84.6+-0.4}{\percent} \\ 
         S5 with ZOH discretization & \SI{74.4+-0.3}{\percent}\\
         S5 with ZOH and $\Delta_k \equiv 1$ & \SI{80.8+-0.1}{\percent} \\
    \bottomrule
    \end{tabular}
    \caption{
        A comparison of our proposed method (Event-SSM) with the Dirac discretized S5 model \eqref{eq:discrete-ssm} (Dirac), S5 with zero-order hold discretization (ZOH), and S5 with all $\Delta_k \equiv 1$, i.e. without parsing information about event timings.
        We report means and standard deviations from 5 runs with random seeds on the SSC dataset.
    }
    \label{tab:ablations}
\end{table}

\section{Discussion}
This work presents a scalable method for the modeling of irregular event-stream data.
Our method addresses the major challenges of event-based processing --- long-range dependencies, asynchronous processing, and parallelization.
The model operates directly on the address event representation of the event-stream and, in contrast to most related works, never uses 2D or 3D convolutions.
The stable state-space model parameterization allows asynchronous recurrent training and inference on very long event-streams of more than a million events, such as those given by event-based vision sensors.
Our ablation study shows that asynchronous event channels require discretization methods that have not yet been used in the machine learning literature.
Furthermore, we observe a clear advantage of integrating exact temporal information compared to binning events into frames for the larger audio processing task SSC.
This result also highlights the need to establish high-quality, large-scale datasets of event-streams 
for challenging machine learning tasks, that will allow us to carefully scrutinize the advantages and disadvantages of event-based machine learning.    

Although our model and all its parts were carefully designed to most effectively operate directly with events, there is no explicit event-generating mechanism in the neural network itself. 
The event-based processing is solely driven by external events. 
On the one hand, this result breaks with the commonly held belief that event data is most effectively processed with event-based neural networks. 
On the other hand, this observation provides an interesting direction for future work to explore how the properties that allow our model to scale to long event-streams can be joined with the efficient processing paradigms of event-based networks.

\begin{acks}
    The authors thank Matthias Jobst for early discussions on the project as well as Jamie Lohoff, Erika Covi and Matthias Jobst for their valuable feedback on the manuscript.
    Mark Sch\"one is supported with funds from Bosch-Forschungsstiftung im Stifterverband.
        David Kappel is funded by the German Federal Ministry for Economic Affairs and Climate Action (BMWK) project ESCADE (01MN23004A).
        Christian Mayr is affiliated to German Research Foundation (DFG, Deutsche Forschungsgemeinschaft) as part of Germany’s Excellence Strategy – EXC 2050/1 – Project ID 390696704 – Cluster of Excellence “Centre for Tactile Internet with Human-in-the-Loop” (CeTI) of Technische Universität Dresden.
        Neeraj Mohan Sushma is funded by the German Federal Ministry of Education and Research (BMBF) project EVENTS (16ME0733).
        This work was partially funded by the German Federal Ministry of Education and Research (BMBF) and the free state of Saxony within the ScaDS.AI center of excellence for AI research.
\end{acks}

\printbibliography

\end{document}